\documentclass[nohyperref]{article}

\usepackage{microtype}
\usepackage{graphicx}
\usepackage{subcaption}
\usepackage{booktabs} 

\usepackage{hyperref}

\usepackage{arxiv}
\usepackage{natbib}

\usepackage{amsmath}
\usepackage{amssymb}
\usepackage{mathtools}
\usepackage{amsthm}

\usepackage[capitalize,noabbrev]{cleveref}

\theoremstyle{plain}
\newtheorem{theorem}{Theorem}[section]

\theoremstyle{definition}

\newtheorem{assumption}[theorem]{Assumption}
\theoremstyle{remark}

\usepackage[textsize=tiny]{todonotes}

\usepackage{enumitem}
\usepackage{xcolor}         
\usepackage{amsmath}
\usepackage[overload]{empheq}
\usepackage{algorithm}
\usepackage{algpseudocode}
\usepackage{placeins}
\usepackage{comment}
\usepackage{bbm}
\usepackage{makecell}

\DeclareMathOperator*{\argmin}{arg\,min}

\date{}
\title{Explicit Regularization via Regularizer Mirror Descent} 

\author{
  Navid Azizan\\
  Massachusetts Institute of Technology\\
  \texttt{azizan@mit.edu} \\
   \And
  Sahin Lale {\normalfont and} Babak Hassibi \\
  California Institute of Technology\\
  \texttt{\{alale,hassibi\}@caltech.edu} \\
}

\begin{document}
\maketitle

\begin{abstract}
  Despite perfectly interpolating the training data, deep neural networks (DNNs) can often generalize fairly well, in part due to the ``implicit regularization'' induced by the learning algorithm. Nonetheless, various forms of regularization, such as ``explicit regularization'' (via weight decay), are often used to avoid overfitting, especially when the data is corrupted. There are several challenges with explicit regularization, most notably unclear convergence properties. Inspired by convergence properties of stochastic mirror descent (SMD) algorithms, we propose a new method for training DNNs with regularization, called regularizer mirror descent (RMD). In highly overparameterized DNNs, SMD simultaneously interpolates the training data and minimizes a certain potential function of the weights.  RMD starts with a standard cost which is the sum of the training loss and a convex regularizer of the weights. Reinterpreting this cost as the potential of an ``augmented'' overparameterized network and applying SMD yields RMD. As a result, RMD inherits the properties of SMD and provably converges to a point ``close'' to the minimizer of this cost. RMD is computationally comparable to stochastic gradient descent (SGD) and weight decay, and is parallelizable in the same manner. Our experimental results on training sets with various levels of corruption suggest that the generalization performance of RMD is remarkably robust and significantly better than both SGD and weight decay, which implicitly and explicitly regularize the $\ell_2$ norm of the weights. RMD can also be used to regularize the weights to a desired weight vector, which is particularly relevant for continual learning.
\end{abstract}

\section{Introduction}

\subsection{Motivation}
Today's deep neural networks are typically highly overparameterized and often have a large enough capacity to easily overfit the training data to zero training error \citep{zhang2016understanding}. Furthermore, it is now widely recognized that such networks can still generalize well despite (over)fitting \citep{bartlett2020benign,belkin2018overfitting,belkin2019reconciling,nakkiran2021deep,bartlett2021deep}, which is, in part, due to the ``implicit regularization'' \citep{gunasekar2018characterizing,azizan2019stochastic,neyshabur2014search,boffi2021implicit} property of the optimization algorithms such as stochastic gradient descent (SGD) or its variants. However, in many cases, especially when the training data is known to include corrupted samples, it is still highly desirable to avoid overfitting the training data through some form of regularization \citep{goodfellow2016regularization,kukavcka2017regularization}. This can be done through, e.g., early stopping, or explicit regularization of the network parameters via weight decay. However, the main challenge with these approaches is that their convergence properties are in many cases unknown and they typically do not come with performance guarantees.


\subsection{Contributions}
The contributions of the paper are as follows.
\begin{enumerate}[wide, labelindent=2pt,label=\textbf{\arabic*})]

    \item We propose a new method for training DNNs with regularization, called regularizer mirror descent (RMD), which allows for choosing any  desired convex regularizer of the weights. 
    RMD leverages the implicit regularization properties of the stochastic mirror descent (SMD) algorithm. It does so by reinterpreting the explicit cost (the sum of the training loss and convex regularizer) as the potential function of an ``augmented'' network. SMD applied to this augmented network and cost results in RMD.
    \item Due to the connection to SMD, contrary to most existing explicit regularization methods, RMD comes with \emph{convergence guarantees}. For highly overparameterized models, it provably converges to a point ``close'' to the minimizer of the cost.
    \item RMD is \emph{computationally and memory-wise efficient}. It imposes virtually no additional overhead compared to standard SGD, and can run in mini-batches and/or be distributed in the same manner.
    \item  We evaluate the performance of RMD using a ResNet-18 neural network architecture on the CIFAR-10 dataset with various levels of corruption.  The results show that the generalization performance of RMD is \emph{remarkably robust to data corruptions} and significantly better than both the standard SGD, which implicitly regularizes the $\ell_2$ norm of the weights, as well as weight decay, which explicitly does so. 
    Further, unlike other explicit regularization methods, e.g., weight decay, the generalization performance of RMD is very consistent and not sensitive to the regularization parameter.
    \item An extension of the convex regularizer can be used to guarantee the closeness of the weights to a desired weight vector with a desired notion of distance. This makes RMD particularly relevant for \emph{continual learning}.
\end{enumerate}
Therefore, we believe that RMD provides a very viable alternative to the existing explicit regularization approaches.

\subsection{Related Work}
There exist a multitude of regularization techniques that are used in conjunction with the training procedures of DNNs. See, e.g., \cite{goodfellow2016regularization,kukavcka2017regularization} for a survey. While it is impossible to discuss every work in the literature, the techniques can be broadly divided into the following categories based on how they are performed: (i) via \emph{data augmentation}, such as mixup \citep{zhang2018mixup}, (ii) via the \emph{network architecture}, such as dropout \citep{hinton2012improving}, and (iii) via the \emph{optimization algorithm}, such as early stopping \citep{li2020gradient,yao2007early,molinari2021iterative}, among others.

Our focus in this work is on explicit regularization, which is done through adding a regularization term to the cost. Therefore, the most closely comparable approach is weight decay \citep{zhang2018three}, which adds an $\ell_2$-norm regularizer to the objective. However, our method is much more general, as it can handle any desired strictly-convex regularizer.

As mentioned earlier, our algorithm for solving the explicitly-regularized problem leverages the ``implicit regularization'' behavior of a family of optimization algorithms called stochastic mirror descent \citep{azizan2019stochastic,azizan2020stochasticTNNLS}. We discuss this further in Section~\ref{sec:background_implicitreg}.


The rest of the paper is organized as follows. We review some preliminaries about explicit and implicit regularization in Section~\ref{sec:background}. We present the main RMD algorithm and its various special cases in Section~\ref{sec:RMD}. In Section~\ref{sec:experiments}, we perform an experimental evaluation of RMD and demonstrate its generalization performance. In Section~\ref{sec:continual}, we show that RMD can be readily used for regularizing the weights to be close to any desired weight vector, which is particularly important for continual learning. We present the convergence guarantees of RMD in Section~\ref{sec:guarantee}, and finally conclude in Section~\ref{sec:conclusion}.


\section{Background}\label{sec:background}
We review some background about stochastic gradient methods and different forms of regularization.

\subsection{Stochastic Gradient Descent}
Let $L_i(w)$  denote the loss on the data point $i$ for a weight vector $w\in\mathbb{R}^p$. For a training set consisting of $n$ data points, the total loss is $\sum_{i=1}^n L_i(w)$, which is typically attempted to be minimized by stochastic gradient descent \citep{robbins1951stochastic} or one of its variants (such as mini-batch, distributed, adaptive, with momentum, etc.). Denoting the model parameters at the $t$-th time step by $w_t\in\mathbb{R}^p$ and the index of the chosen data point by $i$, the update rule of SGD can be simply written as
\begin{equation}\label{eq:SGD}
w_t=w_{t-1}-\eta\nabla L_i(w_{t-1}),\quad t\geq 1,
\end{equation}
where $\eta$ is the so-called learning rate, $w_0$ is the initialization, and $\nabla L_i(\cdot)$ is the gradient of the loss. 
When trained with SGD, typical deep neural networks (which have many more parameters than the number of data points) often achieve (near) zero training error \citep{zhang2016understanding}, or, in other words, ``interpolate'' the training data \citep{mababe2017}.

\subsection{Explicit Regularization}
As mentioned earlier, it is often desirable to avoid (over)fitting the training data to zero error, e.g., when the data has some corrupted labels. In such scenarios, it is beneficial to augment the loss function with a (convex and differentiable) regularizer $\psi:\mathbb{R}^p\to\mathbb{R}$, and consider
\begin{empheq}[box=\fbox]{equation}\label{opt_general_noBreg}
\min_w\ \lambda\sum_{i=1}^nL_i(w)+\psi(w),
\end{empheq}
where $\lambda\geq 0$ is a hyper-parameter that controls the strength of regularization relative to the loss function. A simple and common choice of regularizer is  $\psi(w)=\frac12\|w\|^2$. In this case, when SGD is applied to \eqref{opt_general_noBreg} it is commonly referred to as weight decay.
Note that the bigger $\lambda$ is, the more effort in the optimization is spent on minimizing $\sum_{i=1}^nL_i(w)$. Since the losses $L_i(\cdot)$ are non-negative, the lowest these terms can get is zero, and thus, for $\lambda\to\infty$, the problem would be equivalent to the following:
\begin{equation}\label{eq:min_psi}
\begin{aligned}
& \underset{w}{\text{min}}
& & \psi(w)\\
& \text{s.t.}
& & L_i(w)=0,\quad i=1,\dots,n .
\end{aligned}
\end{equation}

\subsection{Implicit Regularization}\label{sec:background_implicitreg}

Recently, it has been noted in several papers that, even \emph{without} imposing any explicit regularization in the objective, i.e., by optimizing only the loss function $\sum_{i=1}^n L_i(w)$, there is still an implicit regularization induced by the optimization algorithm used for training \citep{gunasekar2018characterizing,gunasekar2018implicit,azizan2019stochastic}. Namely, with sufficiently small step size, SGD tends to converge to interpolating solutions with minimum $\ell_2$ norm \citep{gunasekar2018characterizing,poggio2020theoretical}, i.e.,\footnote{See Section~\ref{sec:guarantee} for a more precise statement.}
\begin{equation*}
\begin{aligned}
& \underset{w}{\text{min}}
& & \|w\|_2\\
& \text{s.t.}
& & L_i(w)=0,\quad i=1,\dots,n .
\end{aligned}
\end{equation*}

More generally, it has been shown \citep{azizan2019stochastic,azizan2019stochasticCDC,azizan2020stochasticTNNLS} that SMD, whose update rule is defined for a differentiable strictly-convex ``potential function'' $\psi(\cdot)$ as
\begin{equation}\label{eq:SMD}
\nabla\psi(w_t) =\nabla\psi(w_{t-1})-\eta\nabla L_i(w_{t-1}) ,
\end{equation}
with proper initialization ($w_0\approx0$)\footnote{For practical reasons, one might not be able to set the initial weight vector exactly to zero. However, it can be initialized randomly around zero, which is common practice, and that would be of no major consequence.} and sufficiently small learning rate converges to the solution of\footnote{See Section~\ref{sec:guarantee} and Theorem~\ref{thm} for a more precise statement.}
\begin{empheq}[box=\fbox]{equation}\label{opt_implicit}
\begin{aligned}
& \underset{w}{\text{min}}
& & \psi(w)\\
& \text{s.t.}
& & L_i(w)=0,\quad i=1,\dots,n .
\end{aligned}
\end{empheq}
Note that this is equivalent to the case of explicit regularization with $\lambda\to\infty$, i.e., problem~\eqref{eq:min_psi}.

\section{Proposed Method: Regularizer Mirror Descent (RMD)}\label{sec:RMD}


When it is undesirable to reach zero training error, e.g., due to the presence of corrupted samples in the data, one cannot rely on the implicit bias of the optimization algorithm to avoid overfitting. That is because these algorithms would interpolate the corrupted data as well. This suggests using explicit regularization as in (\ref{opt_general_noBreg}). Unfortunately, standard explicit regularization methods, such as weight decay, which is simply employing SGD to (\ref{opt_general_noBreg}), do not come with convergence guarantees. Here, we propose a new algorithm, called Regularizer Mirror Descent (RMD), which, under appropriate conditions, provably regularizes the weights for any desired differentiable strictly-convex regularizer $\psi(\cdot)$. In other words, RMD converges to a weight vector close to the minimizer of (\ref{opt_general_noBreg}).

We are interested in solving the explicitly-regularized optimization problem~\eqref{opt_general_noBreg}. Let us define an auxiliary variable $z\in\mathbb{R}^n$ with elements $z[1],\dots,z[n]$. The optimization problem~\eqref{opt_general_noBreg} can be transformed into the following form:
\begin{equation}\label{opt_RMD_noBreg}
\begin{aligned}
& \underset{w,z}{\text{min}}
& &  {\lambda}\sum_{i=1}^n \frac{z^2[i]}{2} +\psi(w)\\
& \text{s.t.}
& & z[i]=\sqrt{2L_i(w)},\quad i=1,\dots,n .
\end{aligned}
\end{equation}
The objective of this optimization problem is a strictly-convex function 
\begin{equation*} \hat\psi\left(w, z\right)=\psi(w)+\frac{\lambda}{2}\|z\|^2,
\end{equation*}
and there are $n$ equality constraints. We can therefore think of an ``augmented'' network with two sets of weights, $w$ and $z$. To enforce the constraints $z[i]=\sqrt{2L_i(w)}$, we can define a ``constraint-enforcing'' loss $\hat\ell\left(z[i]-\sqrt{2L_i(w)}\right)$, where $\hat\ell(\cdot)$ is a differentiable convex function with a unique root at $0$ (e.g., the square loss $\hat\ell(\cdot) = \frac{(\cdot)^2}{2}$). Thus, (\ref{opt_RMD_noBreg}) can be rewritten as
\begin{empheq}[box=\fbox]{equation}\label{opt_explicit_aug}
\begin{aligned}
& \underset{w,z}{\text{min}}
& & \hat\psi(w,z)\\
& \text{s.t.}
& & \hat\ell\left(z[i]-\sqrt{2L_i(w)}\right) = 0,\quad i=1,\dots,n .
\end{aligned}
\end{empheq}
Note that (\ref{opt_explicit_aug}) is similar to the implicitly-regularized optimization problem~\eqref{opt_implicit}, which can be solved via SMD. To do so, we need to follow (\ref{eq:SMD}) and compute the gradients of the potential $\hat\psi(\cdot,\cdot)$, as well as the loss $\hat\ell\left(z[i]-\sqrt{2L_i(w)}\right)$, with respect to $w$ and $z$. We omit the details of this straightforward calculation and simply state the result, which we call the RMD algorithm. 

At time $t$, when the $i$-th training sample is chosen for updating the model, the update rule of RMD can be written as follows:
\begin{align}\label{eq:SMD-z}
\nabla\psi(w_t) &=\nabla\psi(w_{t-1})+\frac{c_{t,i}}{\sqrt{2L_{i}(w_{t-1})}}\nabla L_i(w_{t-1}) ,\notag\\
z_t[i] &= z_{t-1}[i]-\frac{c_{t,i}}{\lambda}, \notag\\
z_t[j] &= z_{t-1}[j],\quad \forall j\neq i,
\end{align}
where 
$c_{t,i}=\eta \hat\ell'\left(z_{t-1}[i] - \sqrt{2L_{i}(w_{t-1})}\right)$, $\hat\ell'(\cdot)$ is the derivative of the constraint-enforcing loss function, and the variables are initialized with $w_0=0$ and $z_0=0$. Note that because of the strict convexity of the regularizer $\psi(\cdot)$, its gradient $\nabla\psi(\cdot)$ is an invertible function, and the above update rule is well-defined. Algorithm~\ref{alg} summarizes the procedure. As will be shown in Section \ref{sec:guarantee}, under suitable conditions, RMD provably solves the optimization problem~\eqref{opt_general_noBreg}.

One can choose the constraint-enforcing loss as $\hat\ell(\cdot) = \frac{(\cdot)^2}{2}$, which implies $\hat\ell'(\cdot) = (\cdot)$, to simply obtain the same update rule as in \eqref{eq:SMD-z} with $c_{t,i} = \eta \left(z_{t-1}[i] - \sqrt{2L_{i}(w_{t-1})}\right)$.

\begin{algorithm}[t]
\caption{Regularizer Mirror Descent (RMD)}\label{alg}
\begin{algorithmic}[1]
\Require{$\lambda, \eta, w_0$}
\State {\bfseries Initialization:} $w\gets w_0$, $z\gets 0$
\Repeat
\State Take a data point $i$ 
\State $c\gets\eta \hat\ell'\left(z[i] - \sqrt{2L_{i}(w)}\right)$
\State $w\gets \nabla\psi^{-1} \left(\nabla\psi(w)+\frac{c}{\sqrt{2L_{i}(w)}}\nabla L_i(w)\right)$
\State $z[i]\gets z[i]-\frac{c}{\lambda}$
\Until{convergence}
\State \Return $w$
\end{algorithmic}
\end{algorithm}

\subsection{Special Case: $q$-norm Potential}
An important special case of RMD is when the potential function $\psi(\cdot)$ is chosen to be the $\ell_q$ norm, i.e., $\psi(w)=\frac{1}{q}\|w\|_q^q = \frac{1}{q} \sum_{k=1}^p |w[k]|^q $, for a real number $q>1$. Let the current gradient be denoted by $g:=\nabla L_i(w_{t-1})$. In this case, the update rule can be written as
\begin{align}\label{eq:q-SMD-z}
w_{t}[k] &= \big|  \xi_{t,i} \big| ^{\frac{1}{q-1}}
\operatorname{sign}\big( \xi_{t,i} \big),\quad \forall k
\notag\\
 z_t[i] &=  z_{t-1}[i]-\frac{c_{t,i}}{\lambda}, \notag\\
z_t[j] &= z_{t-1}[j],\qquad \forall j\neq i,
\end{align}
for $\xi_{t,i} = | w_{t-1}[k] | ^ {q-1} \operatorname{sign}(w_{t-1}[k]) +\frac{c_{t,i}}{\sqrt{2L_{i}(w_{t-1})}}g[k] $, where $w_t[k]$ denotes the $k$-th element of $w_t$ (the weight vector at time $t$) and $g[k]$ is the $k$-th element of the current gradient $g$. Note that for this choice of potential function, the update rule is \emph{separable}, in the sense that the update for the $k$-th element of the weight vector requires only the $k$-th element of the weight and gradient vectors. This allows for efficient parallel implementation of the algorithm, which is crucial for large-scale tasks.



Even among the family of $q$-norm RMD algorithms, there can be a wide range of regularization effects for different values of $q$. Some important examples are as follows:

\textbf{$\mathbf{\ell_1}$ norm} regularization promotes sparsity in the weights. Sparsity is often desirable for reducing the storage and/or computational load, given the massive size of state-of-the-art DNNs. However, since the  $\ell_1$-norm is neither differentiable nor strictly convex, one may use $\psi(w) = \frac{1}{1+\epsilon} \| w \|_{1 + \epsilon}^{1+\epsilon}$ for some small $\epsilon>0$ \citep{azizan2019characterization}. 

\textbf{$\mathbf{\ell_\infty}$ norm} regularization promotes bounded and small range of weights. With this choice of potential, the weights tend to concentrate around a small interval. This is often desirable in various implementations of neural networks since it provides a small dynamic range for quantization of weights, which reduces the production cost and computational complexity. However, since $\ell_\infty$ is, again, not differentiable, one can choose a large value for $q$ and use $\psi(w) = \frac{1}{q}\|w \|_{q}^{q}$ to achieve the desirable regularization effect of $\ell_\infty$ norm ($q=10$ is used in \cite{azizan2020stochasticTNNLS}).

\textbf{$\mathbf{\ell_2}$ norm} still promotes small weights, similar to $\mathbf{\ell_1}$ norm, but to a lesser extent. The update rule is
\begin{align}\label{eq:q-SMD-SGD-with_l}
  w_{t}[k] &=  w_{t-1}[k]+\frac{c_{t,i}}{\sqrt{2L_{i}(w_{t-1})}}g[k], \quad \forall k \notag\\
z_t[i] &= z_{t-1}[i]-\frac{c_{t,i}}{\lambda}, \notag\\
z_t[j] &= z_{t-1}[j],\qquad \forall j\neq i.
\end{align}

\subsection{Special Case: Negative Entropy Potential}
One can choose the potential function $\psi(\cdot)$ to be the negative entropy, i.e., $\psi(w)= \sum_{k=1}^p w[k] \log(w[k])$. For this particular choice, the Bregman divergence reduces to the 
Kullback–Leibler divergence. Let the current gradient be denoted by $g:=\nabla L_i(w_{t-1})$. The update rule would be
\begin{align}\label{eq:entropy-SMD-z}
w_{t}[k] &=  w_{t-1}[k] \exp\left(\frac{c_{t,i}}{\sqrt{2L_{i}(w_{t-1})}}g[k]\right)\quad \forall k\notag\\
 z_t[i] &=  z_{t-1}[i]-\frac{c_{t,i}}{\lambda}, \notag\\
z_t[j] &= z_{t-1}[j],\qquad \forall j\neq i,
\end{align}
This update rule requires the weights to be positive. 

\section{Experimental Results}\label{sec:experiments}

As mentioned in the introduction, there are many ways to regularize DNNs and improve their generalization performance, including methods that perform data augmentation, a change to the network architecture, or early stopping. However, since in this paper we are concerned with the effect of the learning algorithm, we will focus on comparing RMD with the standard SGD (which induces implicit regularization) and the standard weight decay (which attempts to explicitly regularize the $\ell_2$ norm of the weights). No doubt the results can be improved by employing the aforementioned methods with these algorithms, but we leave that study for the future since it will not allow us to isolate the effect of the algorithm. 

As we shall momentarily see, the results indicate that RMD outperforms both alternatives by a significant margin, thus making it a viable option for explicit regularization.

\subsection{Setup}\label{sec:experiments_setup}

\textbf{Dataset.} To test the performance of different regularization methods in avoiding overfitting, we need a training set that does not consist entirely of clean data. We, therefore, took the popular CIFAR-10 dataset \citep{krizhevsky2009learning}, which has $10$ classes and $n=50,000$ training data points, and considered corrupting different fractions of the data. In the first scenario, we corrupted $25\%$ of the data points, by assigning them a random label. Since, for each of those images, there is a $9/10$ chance of being assigned a wrong label, roughly $9/10\times25\% = 22.5\%$ of the training data had incorrect labels. In the second scenario, we randomly flipped $10\%$ of the labels, resulting in roughly $9\%$ incorrect labels. For the sake of comparison, in the third scenario, we considered the uncorrupted data set itself. 

\textbf{Network Architecture.} We used a standard ResNet-18 \citep{he2016deep} deep neural network, which is commonly used for the CIFAR-10 dataset. The network has $18$ layers, and around $11$ million parameters. Thus, it qualifies as highly overparameterized. We did not make any changes to the network.

\textbf{Algorithms.} We use three different algorithms for optimization/regularization.
\begin{enumerate}
\item Standard SGD (implicit regularization): First, we train the network with the standard (mini-batch) SGD. While there is no explicit regularization, this is still known to induce an implicit regularization, as discussed in Section~\ref{sec:background_implicitreg}.

\item Weight decay (explicit regularization): We next train the network with an $\ell_2$-norm regularization, through weight decay. We ran weight decay with a wide range of regularization parameters, $\lambda$.

\item RMD (explicit regularization): Finally, we train the network with RMD, which is provably regularizing with an $\ell_2$ norm. For RMD we also ran the algorithm for a wide range of regularization parameters, $\lambda$.
\end{enumerate}
In all three cases, we train in mini batches---mini-batch RMD is summarized in Algorithm~\ref{alg:minibatch} in the Appendix.

\begin{figure}[t]
\centering
\begin{subfigure}{.445\textwidth}
  \centering
  \includegraphics[width=0.90\linewidth]{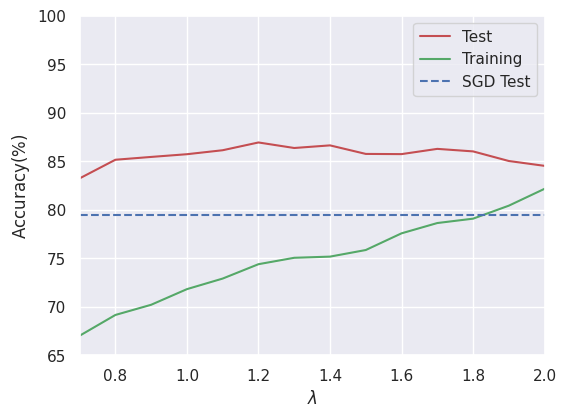}  
  \caption{RMD}
  \label{fig:RMD25}
\end{subfigure}
\begin{subfigure}{.445\textwidth}
  \centering
  \includegraphics[width=0.90\linewidth]{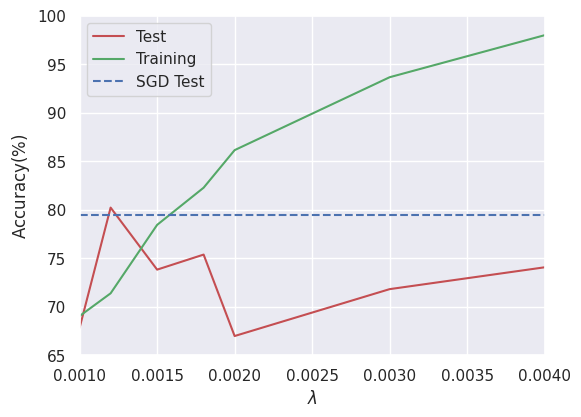}  
  \caption{Weight Decay}
  \label{fig:DECAY25}
\end{subfigure}
\caption{{\small $25\%$ corruption of the training set.}}

\label{fig:25corruption}
\end{figure}

\subsection{Results}

\begin{figure}[t]
\centering
\begin{subfigure}{.445\textwidth}
  \centering
  \includegraphics[width=0.90\linewidth]{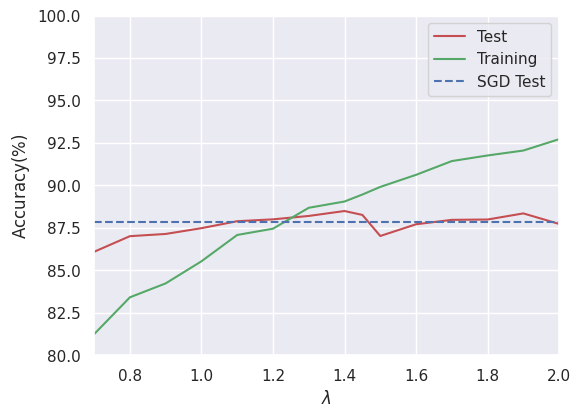}
  \caption{RMD}
  \label{fig:RMD10}
\end{subfigure}
\begin{subfigure}{.445\textwidth}
  \centering
 \includegraphics[width=0.90\linewidth]{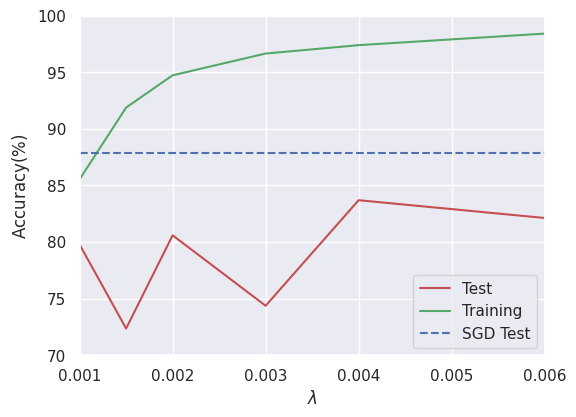}
  \caption{Weight Decay}
  \label{fig:DECAY10}
\end{subfigure}
\caption{{\small $10\%$ corruption of the training set.}}
\label{fig:10corruption}
\end{figure}
\begin{figure}[t]
\centering
\begin{subfigure}{.445\textwidth}
  \centering
 \includegraphics[width=0.90\linewidth]{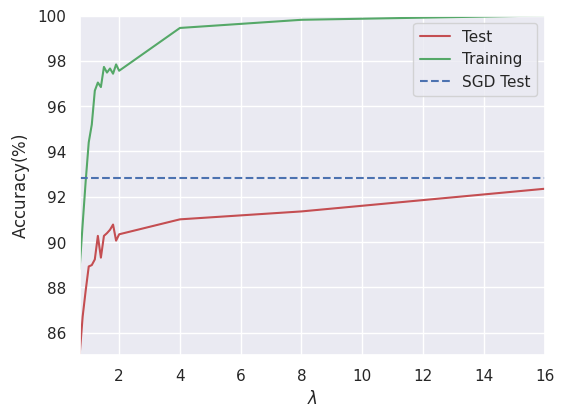}
  \caption{RMD}
  \label{fig:RMD0}
\end{subfigure}
\begin{subfigure}{.445\textwidth}
  \centering
\includegraphics[width=0.90\linewidth]{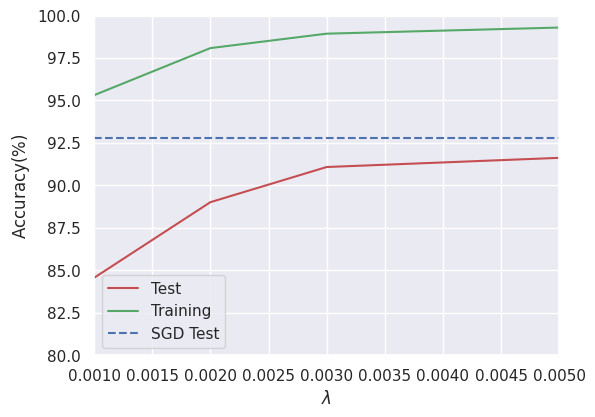}
  \caption{Weight Decay }
  \label{fig:DECAY0}
\end{subfigure}
\caption{{\small Uncorrupted training set.}}
\label{fig:0corruption}
\end{figure}

The training and test accuracies for all three methods are given in Figs.~\ref{fig:25corruption}-\ref{fig:0corruption}. Fig.~\ref{fig:25corruption} shows the results when the training data is corrupted by $25\%$, Fig.~\ref{fig:10corruption} when it is corrupted by $10\%$, and Fig.~\ref{fig:0corruption} when it is uncorrupted. 

As expected, because the network is highly overparameterized, in all cases, SGD interpolates the training data and achieves almost $100\%$ training accuracy. 

As seen in Fig.~\ref{fig:25corruption}, at $25\%$ data corruption SGD achieves $80\%$ test accuracy. For RMD, as $\lambda$ varies from 0.7 to 2.0, the training accuracy increases from $67\%$ to $82\%$ (this increase is expected since RMD should interpolate the training data as $\lambda\rightarrow\infty$). However, the test accuracy remains generally constant around $85\%$, with a peak of $87\%$. This is significantly better than the generalization performance of SGD. For weight decay, as $\lambda$ increases from 0.001 to 0.004, the training accuracy increases from $70\%$ to $98\%$ (implying that there is no need to increase $\lambda$ beyond 0.004). The test accuracy, on the other hand, is rather erratic and varies from a low of $67\%$ to a peak of $80\%$. 

As seen in Fig.~\ref{fig:10corruption}, at $10\%$ data corruption SGD achieves $87.5\%$ test accuracy. For RMD, as $\lambda$ varies from 0.7 to 2.0, the training accuracy increases from $82\%$ to $92.5\%$. The test accuracy remains generally constant around and the peak of $88.5\%$ is only marginally better than SGD. For weight decay, the training accuracy increases from $86\%$ to $99\%$, while the test accuracy is erratic and peaks only at $80\%$. 

Finally, for the sake of comparison, we show the results for the uncorrupted training data in Fig.~\ref{fig:0corruption}. As expected, since the data is uncorrupted, interpolating the data makes sense and SGD has the best test accuracy. Both RMD and weight decay approaches have higher test accuracy as $\lambda$ increases, with RMD having superior performance. 

We should also mention that we have run experiments with $40\%$ corruption in the data. Here SGD achieves $70\%$ test accuracy, while RMD achieves a whopping $81.5\%$ test accuracy with only $64\%$ training accuracy. See the Appendix for more details.



    
     
   
    
        

\section{Regularization for Continual Learning}\label{sec:continual}
It is often desirable to regularize the weights to remain close to a particular weight vector. This is particularly useful for continual learning, where one seeks to learn a new task while trying not to ``forget'' the previous task as much as possible \citep{lopez2017gradient,kirkpatrick2017overcoming,farajtabar2020orthogonal}. In this section, we show that our algorithm can be readily used for such settings by initializing $w_0$ to be the desired weight vector and suitably choosing a notion of closeness.

Augmenting the loss function with a regularization term that promotes closeness to some desired weight vector $w^{\text{reg}}$, one can pose the optimization problem as
\begin{equation}
\begin{aligned}
& \underset{w}{\text{min}}
& &  \lambda\sum_{i=1}^n L_i(w) +\|w-w^{\text{reg}}\|^2.
\end{aligned}
\end{equation}

More generally, using a Bregman divergence $D_{\psi}(\cdot,\cdot)$ corresponding to a differentiable strictly-convex potential function $\psi:\mathbb{R}^p\to\mathbb{R}$, one can pose the problem as
\begin{empheq}[box=\fbox]{equation}\label{opt_general}
\begin{aligned}
& \underset{w}{\text{min}}
& &  \lambda\sum_{i=1}^n L_i(w) +D_{\psi}(w,w^{\text{reg}}).
\end{aligned}
\end{empheq}
Note that Bregman divergence is defined as $D_{\psi}(w,w^{\text{reg}}):=\psi(w)-\psi(w^{\text{reg}})-\nabla\psi(w^{\text{reg}})^T (w-w^{\text{reg}})$, is non-negative, and convex in its first argument. Due to strict convexity of $\psi$, we also have $D_\psi(w,w^{\text{reg}}) = 0$ iff $w=w^{\text{reg}}$. For the choice of $\psi(w)=\frac{1}{2}\|w\|^2$, for example, the Bregman divergence reduces to the usual Euclidean distance $D_{\psi}(w,w_0)=\frac{1}{2}\|w-w^{\text{reg}}\|^2$.

Same as in Section~\ref{sec:RMD}, we can define an auxiliary variable $z\in\mathbb{R}^n$, and rewrite the problem as
\begin{equation}\label{opt_RMD_Breg}
\begin{aligned}
& \underset{w,z}{\text{min}}
& &  {\lambda}\sum_{i=1}^n \frac{z^2[i]}{2} +D_{\psi}(w,w^{\text{reg}})\\
& \text{s.t.}
& & z[i]=\sqrt{2L_i(w)},\quad i=1,\dots,n .
\end{aligned}
\end{equation}
It can be easily shown that the objective of this optimization problem is a Bregman divergence, i.e., $D_{\hat\psi}\left(\begin{bmatrix}w\\z\end{bmatrix},\begin{bmatrix}w^{\text{reg}}\\ {0}\end{bmatrix}\right)$, corresponding to a potential function $\hat\psi\left(\begin{bmatrix}w\\z\end{bmatrix}\right)=\psi(w)+\frac{\lambda}{2}\|z\|^2$. As will be discussed in Section~\ref{sec:guarantee}, this is exactly in a form that an SMD algorithm with the choice of potential function $\hat\psi$, initialization $w_0=w^{\text{reg}}$ and $z_0=0$, and a sufficiently small learning rate will solve.
In other words, Algorithm~\ref{alg} with initialization $w_0=w^{\text{reg}}$ provably solves the regularized problem~\eqref{opt_general}.

\section{Convergence Guarantees}\label{sec:guarantee}
In this section, we provide convergence guarantees for RMD under certain assumptions, motivated by the implicit regularization property of stochastic mirror descent, recently established in \cite{azizan2019stochastic,azizan2020stochasticTNNLS}.

Let us denote the training dataset by $\{(x_i,y_i): i=1,\dots,n\}$, where $x_i\in\mathbb{R}^d$ are the inputs, and $y_i\in\mathbb{R}$ are the labels. The output of the model on data point $i$ is denoted by a function $f_i(w):=f(x_i,w)$ of the parameter $w\in\mathbb{R}^p$. The loss on data point $i$ can then be expressed as $L_i(w)=\ell(y_i-f_i(w))$ with $\ell(\cdot)$ being convex and having a global minimum at zero (examples include square loss, Huber loss, etc.). Since we are mainly concerned with highly overparameterized models (the interpolating regime), where $p\!\gg\! n$, there are (infinitely) many parameter vectors $w$ that can perfectly fit the training data points, and we can define
\begin{align*}
\mathcal{W}&=\{w\in\mathbb{R}^p\ |\ f_i(w)=y_i,\ i=1,\dots,n\}  \\ &=\{w\in\mathbb{R}^p\ |\ L_i(w)=0,\ i=1,\dots,n\}.
\end{align*}
Let $w^*\in\mathcal{W}$ denote the interpolating solution that is closest to the initialization $w_0$ in Bregman divergence:
\begin{equation}\label{eq:w^*}
\begin{aligned}
w^*= & \argmin_{w} & &  D_{\psi}(w,w_0)\\
&\quad\text{ s.t.} & & f_i(w)=y_i,\quad i=1,\dots,n .
\end{aligned}
\end{equation}
It has been shown that, for a linear model $f(x_i,w)=x_i^Tw$, and for a sufficiently small learning rate $\eta>0$, the iterates of SMD \eqref{eq:SMD} with potential function $\psi(\cdot)$, initialized at $w_0$, converge to $w^*$ \citep{azizan2019stochastic}.

When initialized at $w_0=\argmin_{w} \psi(w)$ (which is the origin for all norms, for example), the convergence point becomes the minimum-norm interpolating solution, i.e.,
\begin{equation}
\begin{aligned}
w^*= & \argmin_{w} & &  \psi(w)\\
&\quad\text{ s.t.} & & f_i(w)=y_i,\quad i=1,\dots,n .
\end{aligned}
\end{equation}

While for nonlinear models, the iterates of SMD do not necessarily converge to $w^*$, it has been shown that for highly-overparameterized models, under certain conditions, this still holds in an approximate sense \citep{azizan2020stochasticTNNLS}. In other words, the iterates converge to an interpolating solution $w_{\infty}\in\mathcal{W}$ which is ``close'' to $w^*$. More formally, the result from \cite{azizan2020stochasticTNNLS} along with its assumptions can be stated as follows.

Let us define
$D_{L_i}(w,w'):=L_i(w)-L_i(w')-\nabla L_i(w')^T(w-w'),$
which is defined in a similar way to a Bregman divergence for the loss function. The difference though is that, unlike the potential function of the Bregman divergence, due to the nonlinearity of $f_i(\cdot)$, the loss function $L_i(\cdot) = \ell(y_i-f_i(\cdot))$ need not be convex (even when $\ell(\cdot)$ is). Further, denote the Hessian of $f_i$ by $H_{f_i}$.\footnote{We refrain from using $\nabla^2 f_i$ for Hessian, which is typically used for Laplacian (divergence of the gradient).}

\begin{assumption}[\citet{azizan2020stochasticTNNLS}]\label{assump1}
Denote the initial point by $w_0$. There exists $w\in\mathcal{W}$ and a region $\mathcal{B}=\{w'\in\mathbb{R}^p\ |\ D_{\psi}(w,w')\leq\epsilon\}$ containing $w_0$, such that $D_{L_i}(w,w')\geq 0, i=1,\dots,n$, for all $w'\in\mathcal{B}$.
\end{assumption}
\begin{assumption}[\citet{azizan2020stochasticTNNLS}]\label{assump2}
Consider the region $\mathcal{B}$ in Assumption~\ref{assump1}. The $f_i(\cdot)$ have bounded gradient and Hessian on the convex hull of $\mathcal{B}$, i.e., $\|\nabla f_i(w')\|\leq \gamma$, and
$\alpha\leq\lambda_{\min}(H_{f_i}(w'))\leq\lambda_{\max}(H_{f_i}(w'))\leq\beta, i=1,\dots,n$, for all $w'\in\mathrm{conv}\ \mathcal{B}$.
\end{assumption}
\begin{theorem}[\citet{azizan2020stochasticTNNLS}]\label{thm}
Consider the set of interpolating solutions $\mathcal{W}=\{w\in\mathbb{R}^p\ |\ f(x_i,w)=y_i,\ i=1,\dots,n\}$, the closest such solution $w^*=\argmin_{w\in\mathcal{W}} D_{\psi}(w,w_0)$, and the SMD iterates given in \eqref{eq:SMD} initialized at $w_0$, where every data point is revisited after some steps. Under Assumptions~\ref{assump1} and \ref{assump2}, for sufficiently small step size, i.e., for any $\eta>0$ for which $\psi(\cdot)-\eta L_i(\cdot)$ is strictly convex on $\mathcal{B}$ for all i, the following holds.
\begin{enumerate}
    \item The iterates converge to $w_{\infty}\in\mathcal{W}$.
    \item $D_{\psi}(w^*,w_{\infty})=o(\epsilon)${.}
\end{enumerate}
\end{theorem}
In a nutshell, Assumption~\ref{assump1} states that the initial point $w_0$ is close to the set of global minima $\mathcal{W}$, which arguably comes for free in highly overparameterized settings \citep{allen2019convergence}
, while Assumption~\ref{assump2} states that the first and second derivatives of the model are \emph{locally} bounded. Motivated by the above result, we now return to RMD and its corresponding optimization problem.

Let us define a learning problem over parameters $\begin{bmatrix}w\\z\end{bmatrix}\in\mathbb{R}^{p+n}$ with $\hat f_i\left(\begin{bmatrix}w\\z\end{bmatrix}\right)=\sqrt{2L_i(w)}-z[i]$, $\hat y_i=0$, and $\hat L_i\left(\begin{bmatrix}w\\z\end{bmatrix}\right)=\hat\ell\left(\hat y_i-\hat f_i\left(\begin{bmatrix}w\\z\end{bmatrix}\right)\right)=\hat\ell\left(z[i]\!-\!\sqrt{2L_i(w)}\right)$ for $i\!=\!0,\dots,n$. Note that in this new problem, we now have $p\!+\!n$ parameters and $n$ constraints/data points, and since $p\gg n$, we have $p+n\!\gg\! n$, and we are still in the highly-overparameterized regime (even more so). Thus, we can also define the set of interpolating solutions for the new problem as
\begin{equation}\label{hat W}
\hat{\mathcal{W}}=\left\{\begin{bmatrix}w\\z\end{bmatrix}\in\mathbb{R}^{p+n}\ |\ \hat f_i(w)\!=\!\hat y_i,\ i=1,\dots,n\right\}.
\end{equation}

Let us define a potential function $\hat\psi\left(\begin{bmatrix}w\\z\end{bmatrix}\right)=\psi(w)+\frac{\lambda}{2}\|z\|^2$ and a corresponding SMD
\begin{equation*}
\nabla\hat\psi\left(\begin{bmatrix}w_t\\z_t\end{bmatrix}\right) \!=\! \nabla\hat\psi\left(\begin{bmatrix}w_{t-1}\\z_{t-1}\end{bmatrix}\right)\!-\!\eta\nabla \hat L_i\left(\begin{bmatrix}w_{t-1}\\z_{t-1}\end{bmatrix}\right), 
\end{equation*}
initialized at $\begin{bmatrix}w_0\\0\end{bmatrix}$. It is straightforward to verify that this update rule is equivalent to that of RMD, i.e., \eqref{eq:SMD-z}.

On the other hand, from \eqref{eq:w^*}, we have
\begin{equation}\label{eq:hat w^*}
\begin{aligned}
\hat w^*= & \argmin_{w,z} & &  D_{\hat\psi}\left(\begin{bmatrix}w\\z\end{bmatrix},\begin{bmatrix}w_0\\ {0}\end{bmatrix}\right)\\
&\quad\text{ s.t.} & & \hat f_i\left(\begin{bmatrix}w\\z\end{bmatrix}\right)=\hat y_i,\quad i=1,\dots,n .
\end{aligned}
\end{equation}
Plugging $D_{\hat\psi}\left(\begin{bmatrix}w\\z\end{bmatrix},\begin{bmatrix}w_0\\ {0}\end{bmatrix}\right)=D_{\psi}(w,w_0)+\frac{\lambda}{2}\|z\|^2$ and $\hat f_i\left(\begin{bmatrix}w\\z\end{bmatrix}\right)\!=\!\sqrt{2L_i(w)}\!-\!z[i]$ into \eqref{eq:hat w^*}, it is easy to see that it is equivalent to \eqref{opt_RMD_Breg} for $w_0=w^{\text{reg}}$, and equivalent to \eqref{opt_RMD_noBreg} for $w_0=0$. The formal statement of the theorem follows from a direct application of Theorem~\ref{thm}.

\begin{assumption}\label{hat assump1}
Denote the initial point by $\begin{bmatrix}w_0\\0\end{bmatrix}$. There exists $\begin{bmatrix}w\\z\end{bmatrix}\in\hat{\mathcal{W}}$ and a region $\hat{\mathcal{B}}\!=\!\left\{\begin{bmatrix}w'\\z'\end{bmatrix}\in\mathbb{R}^{p+n}\ \! |\ \! D_{\hat\psi}\left(\begin{bmatrix}w\\z\end{bmatrix},\begin{bmatrix}w'\\z'\end{bmatrix}\right)\!\leq\!\epsilon\right\}$ containing $\begin{bmatrix}w_0\\0\end{bmatrix}$, such that $D_{\hat L_i}\!\left(\begin{bmatrix}w\\z\end{bmatrix},\begin{bmatrix}w'\\z'\end{bmatrix}\right)\!\geq 0, i=1,\dots,n$, for all $ \begin{bmatrix}w'\\z'\end{bmatrix}\in\hat{\mathcal{B}}$.
\end{assumption}
\begin{assumption}\label{hat assump2}
Consider the region $\hat{\mathcal{B}}$ in Assumption~\ref{hat assump1}. The $\hat f_i(\cdot)$ have bounded gradient and Hessian on the convex hull of $\hat{\mathcal{B}}$, i.e., $\left\|\nabla \hat f_i\left(\begin{bmatrix}w'\\z'\end{bmatrix}\right)\right\|\leq \gamma$, and
$\alpha\leq\lambda_{\min}\left(H_{\hat f_i}\left(\begin{bmatrix}w'\\z'\end{bmatrix}\right)\right)\leq\lambda_{\max}\left(H_{\hat f_i}\left(\begin{bmatrix}w'\\z'\end{bmatrix}\right)\right)\leq\beta, i=1,\dots,n$, for all $\begin{bmatrix}w'\\z'\end{bmatrix}\in\mathrm{conv}\ \hat{\mathcal{B}}$.
\end{assumption}
\begin{theorem}\label{hat thm}
Consider the set of interpolating solutions $\hat{\mathcal{W}}$ defined in \eqref{hat W}, the closest such solution $\hat w^*$ defined in \eqref{eq:hat w^*}, and the RMD iterates given in \eqref{eq:SMD-z} initialized at $\begin{bmatrix}w_0\\0\end{bmatrix}$, where every data point is revisited after some steps. Under Assumptions~\ref{hat assump1} and \ref{hat assump2}, for sufficiently small step size, i.e., for any $\eta>0$ for which $\hat\psi(\cdot)-\eta \hat L_i(\cdot)$ is strictly convex on $\hat{\mathcal{B}}$ for all i, the following holds.
\begin{enumerate}
    \item The iterates converge to $\begin{bmatrix}w_{\infty}\\z_{\infty}\end{bmatrix}\in\hat{\mathcal{W}}$.
    \item $D_{\hat\psi}\left(\hat w^*,\begin{bmatrix}w_{\infty}\\z_{\infty}\end{bmatrix}\right)=o(\epsilon)${.}
\end{enumerate}
\end{theorem}
Despite its somewhat complicated look, similar as in Assumption~\ref{assump1}, Assumption~\ref{hat assump1} states the initial point $\begin{bmatrix}w_0\\0\end{bmatrix}$ is close to the (new) $(p+n)$-dimensional manifold $\hat{\mathcal{W}}$, which is reasonable because the new problem is even more overparameterized than the original $p$-dimensional one. Similar as in Assumption~\ref{assump2}, Assumption~\ref{hat assump2} requires the first and second derivatives of the model to be locally bounded.

We should emphasize that while Theorem~\ref{hat thm} states that we converge to the manifold $\hat{\mathcal{W}}$, it does \emph{not} mean that it is fitting the training data points or achieving zero training error. That is because $\hat{\mathcal{W}}\in\mathbb{R}^{p+n}$ is a different (much higher-dimensional) manifold than $\mathcal{W}\in\mathbb{R}^p$, and interpolating it would translate to fitting the constraints defined by the regularized problem.

\section{Conclusion and Outlook}\label{sec:conclusion}
We presented {Regularizer Mirror Descent (RMD)}, a novel efficient algorithm for training DNN with any desired strictly-convex regularizer.  The starting point for RMD is a  standard cost which is the sum of the training loss and a differentiable strictly-convex regularizer of the network weights. For highly-overparameterized models, RMD provably converges to a point ``close'' to the minimizer of this cost. The algorithm can be readily applied to any DNN and enjoys the same parallelization properties as SGD. We demonstrated that RMD is remarkably robust to various levels of label corruption in data, and it outperforms both the implicit regularization induced by SGD and the explicit regularization performed via weight decay, by a wide margin. We further showed that RMD can be used for continual learning, where regularization with respect to a previously-learned weight vector is critical.

Given that RMD enables training any network efficiently with a desired regularizer, it opens up several new avenues for future research. In particular, an extensive experimental study of the effect of different regularizers on different datasets and different architectures would be instrumental to uncovering the role of regularization in modern learning problems.

\bibliography{references}

\begin{thebibliography}{33}
\providecommand{\natexlab}[1]{#1}
\providecommand{\url}[1]{\texttt{#1}}
\expandafter\ifx\csname urlstyle\endcsname\relax
  \providecommand{\doi}[1]{doi: #1}\else
  \providecommand{\doi}{doi: \begingroup \urlstyle{rm}\Url}\fi

\bibitem[Allen-Zhu et~al.(2019)Allen-Zhu, Li, and Song]{allen2019convergence}
Allen-Zhu, Z., Li, Y., and Song, Z.
\newblock A convergence theory for deep learning via over-parameterization.
\newblock In \emph{Proceedings of the 36th International Conference on Machine
  Learning}. PMLR, 2019.

\bibitem[Azizan \& Hassibi(2019{\natexlab{a}})Azizan and
  Hassibi]{azizan2019characterization}
Azizan, N. and Hassibi, B.
\newblock A characterization of stochastic mirror descent algorithms and their
  convergence properties.
\newblock In \emph{IEEE International Conference on Acoustics, Speech and
  Signal Processing (ICASSP)}, 2019{\natexlab{a}}.

\bibitem[Azizan \& Hassibi(2019{\natexlab{b}})Azizan and
  Hassibi]{azizan2019stochastic}
Azizan, N. and Hassibi, B.
\newblock Stochastic gradient/mirror descent: Minimax optimality and implicit
  regularization.
\newblock In \emph{International Conference on Learning Representations
  (ICLR)}, 2019{\natexlab{b}}.

\bibitem[Azizan \& Hassibi(2019{\natexlab{c}})Azizan and
  Hassibi]{azizan2019stochasticCDC}
Azizan, N. and Hassibi, B.
\newblock A stochastic interpretation of stochastic mirror descent:
  Risk-sensitive optimality.
\newblock In \emph{2019 58th IEEE Conference on Decision and Control (CDC)},
  pp.\  3960--3965, 2019{\natexlab{c}}.

\bibitem[Azizan et~al.(2021)Azizan, Lale, and
  Hassibi]{azizan2020stochasticTNNLS}
Azizan, N., Lale, S., and Hassibi, B.
\newblock Stochastic mirror descent on overparameterized nonlinear models.
\newblock \emph{IEEE Transactions on Neural Networks and Learning Systems},
  2021.
\newblock \doi{10.1109/TNNLS.2021.3087480}.

\bibitem[Bartlett et~al.(2020)Bartlett, Long, Lugosi, and
  Tsigler]{bartlett2020benign}
Bartlett, P.~L., Long, P.~M., Lugosi, G., and Tsigler, A.
\newblock Benign overfitting in linear regression.
\newblock \emph{Proceedings of the National Academy of Sciences}, 117\penalty0
  (48):\penalty0 30063--30070, 2020.

\bibitem[Bartlett et~al.(2021)Bartlett, Montanari, and
  Rakhlin]{bartlett2021deep}
Bartlett, P.~L., Montanari, A., and Rakhlin, A.
\newblock Deep learning: a statistical viewpoint.
\newblock \emph{arXiv preprint arXiv:2103.09177}, 2021.

\bibitem[Belkin et~al.(2018)Belkin, Hsu, and Mitra]{belkin2018overfitting}
Belkin, M., Hsu, D.~J., and Mitra, P.
\newblock Overfitting or perfect fitting? risk bounds for classification and
  regression rules that interpolate.
\newblock In \emph{Advances in Neural Information Processing Systems},
  volume~31, 2018.

\bibitem[Belkin et~al.(2019)Belkin, Hsu, Ma, and Mandal]{belkin2019reconciling}
Belkin, M., Hsu, D., Ma, S., and Mandal, S.
\newblock Reconciling modern machine-learning practice and the classical
  bias--variance trade-off.
\newblock \emph{Proceedings of the National Academy of Sciences}, 116\penalty0
  (32):\penalty0 15849--15854, 2019.

\bibitem[Boffi \& Slotine(2021)Boffi and Slotine]{boffi2021implicit}
Boffi, N.~M. and Slotine, J.-J.~E.
\newblock Implicit regularization and momentum algorithms in nonlinearly
  parameterized adaptive control and prediction.
\newblock \emph{Neural Computation}, 33\penalty0 (3):\penalty0 590--673, 2021.

\bibitem[Caruana et~al.(2001)Caruana, Lawrence, and
  Giles]{caruana2001overfitting}
Caruana, R., Lawrence, S., and Giles, L.
\newblock Overfitting in neural nets: Backpropagation, conjugate gradient, and
  early stopping.
\newblock \emph{Advances in neural information processing systems}, pp.\
  402--408, 2001.

\bibitem[Farajtabar et~al.(2020)Farajtabar, Azizan, Mott, and
  Li]{farajtabar2020orthogonal}
Farajtabar, M., Azizan, N., Mott, A., and Li, A.
\newblock Orthogonal gradient descent for continual learning.
\newblock In \emph{International Conference on Artificial Intelligence and
  Statistics}, pp.\  3762--3773. PMLR, 2020.

\bibitem[Goodfellow et~al.(2016)Goodfellow, Bengio, and
  Courville]{goodfellow2016regularization}
Goodfellow, I., Bengio, Y., and Courville, A.
\newblock Regularization for deep learning.
\newblock \emph{Deep learning}, pp.\  216--261, 2016.

\bibitem[Gunasekar et~al.(2018{\natexlab{a}})Gunasekar, Lee, Soudry, and
  Srebro]{gunasekar2018characterizing}
Gunasekar, S., Lee, J., Soudry, D., and Srebro, N.
\newblock Characterizing implicit bias in terms of optimization geometry.
\newblock In \emph{International Conference on Machine Learning}, pp.\
  1827--1836, 2018{\natexlab{a}}.

\bibitem[Gunasekar et~al.(2018{\natexlab{b}})Gunasekar, Lee, Soudry, and
  Srebro]{gunasekar2018implicit}
Gunasekar, S., Lee, J.~D., Soudry, D., and Srebro, N.
\newblock Implicit bias of gradient descent on linear convolutional networks.
\newblock In \emph{Advances in Neural Information Processing Systems},
  volume~31, 2018{\natexlab{b}}.

\bibitem[He et~al.(2016)He, Zhang, Ren, and Sun]{he2016deep}
He, K., Zhang, X., Ren, S., and Sun, J.
\newblock Deep residual learning for image recognition.
\newblock In \emph{Proceedings of the IEEE conference on computer vision and
  pattern recognition}, pp.\  770--778, 2016.

\bibitem[Hinton et~al.(2012)Hinton, Srivastava, Krizhevsky, Sutskever, and
  Salakhutdinov]{hinton2012improving}
Hinton, G.~E., Srivastava, N., Krizhevsky, A., Sutskever, I., and
  Salakhutdinov, R.~R.
\newblock Improving neural networks by preventing co-adaptation of feature
  detectors.
\newblock \emph{arXiv preprint arXiv:1207.0580}, 2012.

\bibitem[Kirkpatrick et~al.(2017)Kirkpatrick, Pascanu, Rabinowitz, Veness,
  Desjardins, Rusu, Milan, Quan, Ramalho, Grabska-Barwinska,
  et~al.]{kirkpatrick2017overcoming}
Kirkpatrick, J., Pascanu, R., Rabinowitz, N., Veness, J., Desjardins, G., Rusu,
  A.~A., Milan, K., Quan, J., Ramalho, T., Grabska-Barwinska, A., et~al.
\newblock Overcoming catastrophic forgetting in neural networks.
\newblock \emph{Proceedings of the national academy of sciences}, 114\penalty0
  (13):\penalty0 3521--3526, 2017.

\bibitem[Krizhevsky \& Hinton(2009)Krizhevsky and
  Hinton]{krizhevsky2009learning}
Krizhevsky, A. and Hinton, G.
\newblock Learning multiple layers of features from tiny images.
\newblock Technical report, Citeseer, 2009.

\bibitem[Kuka{\v{c}}ka et~al.(2017)Kuka{\v{c}}ka, Golkov, and
  Cremers]{kukavcka2017regularization}
Kuka{\v{c}}ka, J., Golkov, V., and Cremers, D.
\newblock Regularization for deep learning: A taxonomy.
\newblock \emph{arXiv preprint arXiv:1710.10686}, 2017.

\bibitem[Li et~al.(2020)Li, Soltanolkotabi, and Oymak]{li2020gradient}
Li, M., Soltanolkotabi, M., and Oymak, S.
\newblock Gradient descent with early stopping is provably robust to label
  noise for overparameterized neural networks.
\newblock In \emph{International conference on artificial intelligence and
  statistics}, pp.\  4313--4324. PMLR, 2020.

\bibitem[Lopez-Paz \& Ranzato(2017)Lopez-Paz and Ranzato]{lopez2017gradient}
Lopez-Paz, D. and Ranzato, M.~A.
\newblock Gradient episodic memory for continual learning.
\newblock In \emph{Advances in Neural Information Processing Systems},
  volume~30, 2017.

\bibitem[Ma et~al.(2018)Ma, Bassily, and Belkin]{mababe2017}
Ma, S., Bassily, R., and Belkin, M.
\newblock The power of interpolation: Understanding the effectiveness of {SGD}
  in modern over-parametrized learning.
\newblock In \emph{Proceedings of the 35th International Conference on Machine
  Learning}, volume~80, pp.\  3325--3334. PMLR, 2018.

\bibitem[Molinari et~al.(2021)Molinari, Massias, Rosasco, and
  Villa]{molinari2021iterative}
Molinari, C., Massias, M., Rosasco, L., and Villa, S.
\newblock Iterative regularization for convex regularizers.
\newblock In \emph{International Conference on Artificial Intelligence and
  Statistics}, pp.\  1684--1692. PMLR, 2021.

\bibitem[Nakkiran et~al.(2021)Nakkiran, Kaplun, Bansal, Yang, Barak, and
  Sutskever]{nakkiran2021deep}
Nakkiran, P., Kaplun, G., Bansal, Y., Yang, T., Barak, B., and Sutskever, I.
\newblock Deep double descent: Where bigger models and more data hurt.
\newblock \emph{Journal of Statistical Mechanics: Theory and Experiment},
  2021\penalty0 (12):\penalty0 124003, 2021.

\bibitem[Neyshabur et~al.(2015)Neyshabur, Tomioka, and
  Srebro]{neyshabur2014search}
Neyshabur, B., Tomioka, R., and Srebro, N.
\newblock In search of the real inductive bias: On the role of implicit
  regularization in deep learning.
\newblock In \emph{ICLR (Workshop)}, 2015.
\newblock URL \url{http://arxiv.org/abs/1412.6614}.

\bibitem[Poggio et~al.(2020)Poggio, Banburski, and Liao]{poggio2020theoretical}
Poggio, T., Banburski, A., and Liao, Q.
\newblock Theoretical issues in deep networks.
\newblock \emph{Proceedings of the National Academy of Sciences}, 117\penalty0
  (48):\penalty0 30039--30045, 2020.

\bibitem[Prechelt(1998)]{prechelt1998early}
Prechelt, L.
\newblock Early stopping-but when?
\newblock In \emph{Neural Networks: Tricks of the trade}, pp.\  55--69.
  Springer, 1998.

\bibitem[Robbins \& Monro(1951)Robbins and Monro]{robbins1951stochastic}
Robbins, H. and Monro, S.
\newblock A stochastic approximation method.
\newblock \emph{The annals of mathematical statistics}, pp.\  400--407, 1951.

\bibitem[Yao et~al.(2007)Yao, Rosasco, and Caponnetto]{yao2007early}
Yao, Y., Rosasco, L., and Caponnetto, A.
\newblock On early stopping in gradient descent learning.
\newblock \emph{Constructive Approximation}, 26\penalty0 (2):\penalty0
  289--315, 2007.

\bibitem[Zhang et~al.(2016)Zhang, Bengio, Hardt, Recht, and
  Vinyals]{zhang2016understanding}
Zhang, C., Bengio, S., Hardt, M., Recht, B., and Vinyals, O.
\newblock Understanding deep learning requires rethinking generalization.
\newblock \emph{arXiv preprint arXiv:1611.03530}, 2016.

\bibitem[Zhang et~al.(2018{\natexlab{a}})Zhang, Wang, Xu, and
  Grosse]{zhang2018three}
Zhang, G., Wang, C., Xu, B., and Grosse, R.
\newblock Three mechanisms of weight decay regularization.
\newblock In \emph{International Conference on Learning Representations},
  2018{\natexlab{a}}.

\bibitem[Zhang et~al.(2018{\natexlab{b}})Zhang, Cisse, Dauphin, and
  Lopez-Paz]{zhang2018mixup}
Zhang, H., Cisse, M., Dauphin, Y.~N., and Lopez-Paz, D.
\newblock mixup: Beyond empirical risk minimization.
\newblock In \emph{International Conference on Learning Representations},
  2018{\natexlab{b}}.

\end{thebibliography}
\bibliographystyle{icml2022}

\newpage
\appendix
\onecolumn

\begin{center}
{\huge Appendix}
\end{center}

In what follows, we first show the reduction of RMD to SMD for $\lambda\to\infty$. Then, in Appendix~\ref{apx:experiment_detail}, we provide further details on the experiments for the purpose of reproducing the results. Finally, in Appendix~\ref{apx:experiment_results}, we provide the complete set of experimental results.

\section{Reduction of RMD to SMD for $\lambda\to\infty$} \label{apx:SMD_reduction}

Here, we show that, for $\lambda\to\infty$, RMD reduces to the standard SMD, which jives with the fact that the optimization problem it solves, i.e., \eqref{opt_general_noBreg}, for $\lambda\to\infty$ reduces to the optimization problem that SMD solves, i.e., \eqref{opt_implicit}.

Note that when $\lambda \to \infty$, the update rule for $z_t$ in \eqref{eq:SMD-z} vanishes, and we have $z_t=0$ for all $t$. Therefore, the update becomes
\begin{equation}\label{eq:SMD-adaptive}
\nabla\psi(w_t) =\nabla\psi(w_{t-1})+\frac{\eta \ell'\left(- \sqrt{2L_{i}(w_{t-1})}\right)}{\sqrt{2L_{i}(w_{t-1})}}\nabla L_i(w_{t-1}) .
\end{equation}

For $\ell(\cdot)=\frac{(\cdot)^2}{2}$, we have $\ell'(\cdot)=(\cdot)$, and the update rule further reduces to 
\begin{equation}
\nabla\psi(w_t) =\nabla\psi(w_{t-1})-\eta\nabla L_i(w_{t-1}),
\end{equation}
which is precisely the update rule for SMD. In Section

\section{Additional Details on the Experiments}\label{apx:experiment_detail}

\textbf{Dataset.} To test the ability of different regularization methods in avoiding overfitting, we need a training set that does not consist entirely of clean data. Thus, we took the popular CIFAR-10 dataset \citep{krizhevsky2009learning}, which has $10$ classes and $n=50,000$ training data points, and created $3$ new datasets by corrupting $10\%, 25\%, 40\%$ of the data points via assigning them random labels. Note that for each of those images, there is a $9/10$ chance of being assigned a wrong label. Therefore, on average, there are about $9\%, 22.5\%$ and $36\%$ incorrect labels in the aforementioned datasets, respectively. To have a standard baseline, we also run our experiments on the standard CIFAR-10 dataset (i.e., $0\%$ corruption). To have a competitive generalization performance while not creating any variation between experiments, we select a random data augmentation (crop + horizontal flip) for each dataset and use that for every experiment on the same dataset. Therefore, exactly the same data points are used in the experiments that are on the same corruption level dataset. No corruption is applied on the test data, i.e., the standard test set of CIFAR-10 is used for evaluating the test accuracies. 

\textbf{Network Architecture.} We train a standard ResNet-18 \citep{he2016deep} deep neural network, as implemented in \url{https://github.com/kuangliu/pytorch-cifar}, which is commonly used for the CIFAR-10 dataset. The network has $18$ layers, and around $11$ million parameters. Thus, it qualifies as highly overparameterized. We do not make any changes to the network, and we use the same structure in every experiment.

\textbf{Algorithms.} We use three different algorithms for optimization/regularization.
\begin{enumerate}
\item Standard SGD (implicit regularization): First, we train the network with the standard (mini-batch) SGD. While there is no explicit regularization, this is still known to induce an implicit regularization, as discussed in Section~\ref{sec:background_implicitreg}.

\item Explicit regularization via weight decay: This time, we train the network with an explicit $\ell_2$-norm regularization, through weight decay.

\item Explicit regularization via RMD: Finally, we train the network with RMD, which is provably regularizing with an $\ell_2$ norm.
\end{enumerate}

\textbf{Mini Batch.} For all three algorithms, we train in mini batches. The size of mini batches that we use is 128, which is a common choice for CIFAR-10. The mini-batch implementation of RMD is summarized in Algorithm \ref{alg:minibatch}.

\textbf{Initialization.} The parameters $w$ and $z$ are initialized randomly around zero.

\begin{algorithm}[h]
\caption{Mini-batch Regularizer Mirror Descent (RMD)}\label{alg:minibatch}
\begin{algorithmic}[1]
\Require{$\lambda, \eta, w_0$}
\State {\bfseries Initialization:} $w\gets w_0$, $z\gets 0$
\Repeat
\State Take a mini batch $B$
\State $\bar L(w)\gets\frac{1}{|B|}\sum_{i\in B}L_i(w)$ 
\State $\nabla\bar L(w)\gets\frac{1}{|B|}\sum_{i\in B}\nabla L_i(w)$ 
\State $\bar{z}\gets\frac{1}{|B|}\sum_{i\in B}z[i]$
\State $c\gets\eta \hat\ell'\left(\bar z - \sqrt{2\bar L(w)}\right)$
\State $w\gets \nabla\psi^{-1} \left(\nabla\psi(w)+\frac{c}{\sqrt{2\bar L(w)}}\nabla \bar L(w)\right)$
\For{$i\in B$}
\State $z[i]\gets z[i]-\frac{c}{\lambda}$
\EndFor
\Until{convergence}
\State \Return $w$
\end{algorithmic}
\end{algorithm}

\textbf{Learning Rate.} We used three different learning rates for each of the algorithms: $0.001$, $0.01$ and $0.1$. Among all, $0.1$ provided the best convergence behavior for the SGD and RMD, whereas $0.01$ worked the best for the explicit regularization via weight decay. The reported results are given for the best-performing learning rate choices.

\textbf{Stopping Criterion.}
In order to determine the stopping point for each of the algorithms, we use the following stopping criteria.
\begin{enumerate}
\item For SGD, we train until the training data is interpolated, i.e., $100\%$ training accuracy, similar as in \cite{azizan2020stochasticTNNLS}. 

\item Note that it is not feasible to determine the stopping criterion based on the training accuracies for the explicit regularization via weight decay or RMD. Therefore, for the explicit regularization via weight decay, we consider the change in the total loss over the training set. We stop the training if the change in the total loss is less than $0.01\%$ over $500$ consecutive epochs. 

\item For RMD, we know that the algorithm eventually interpolates the new manifold $\hat{\mathcal{W}}$, i.e., fits the constraints in \eqref{opt_RMD_noBreg}. Thus, we can use the total change in the constraints, i.e.,
\begin{equation*}
    \sum_{i=1}^n \left|z[i] - \sqrt{2L_i(w)} \right|.
\end{equation*}
and we stop the training if this summation improves less than $0.01\%$ over $500$ consecutive epochs. 
\end{enumerate}

We should emphasize that, given our choices for the setup, \emph{the only difference between the experiments on the same corruption level dataset is the optimization algorithm.}

\newpage
\section{Complete Experimental Results} \label{apx:experiment_results}

In this section, we present the experimental results on $40\%$, $25\%$, $10\%$ corrupted and uncorrupted CIFAR-10 training sets. 

\subsection{Results on $40\%$ Corruption of the Training Set}

\begin{figure}[h]
\centering
\begin{subfigure}{.483\textwidth}
  \centering
  \includegraphics[width=0.90\linewidth]{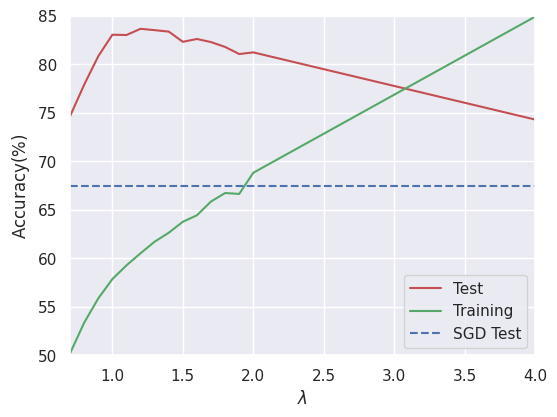}
  \caption{RMD}
  \label{fig:RMD40}
\end{subfigure}
\begin{subfigure}{.483\textwidth}
  \centering
 \includegraphics[width=0.90\linewidth]{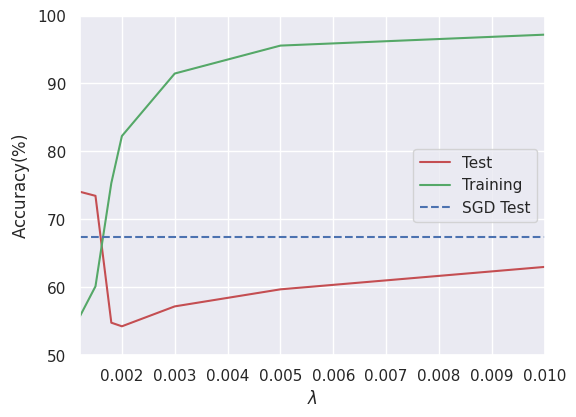}
  \caption{Weight Decay}
  \label{fig:DECAY40}
\end{subfigure}
\caption{{\small $40\%$ corruption of the training set.}}
\vspace{-1em}
\label{fig:40corruption}
\end{figure}

The training and test accuracies for the three algorithms with various values of $\lambda$ are given in Table \ref{table:apx_40} and visualized in Figure \ref{fig:40corruption}. At $40\%$ data corruption, SGD interpolates the training data due to high overparameterization and achieves $100\%$ training accuracy. However, this interpolation results in overfitting and thus $67.41\%$ test accuracy.

Varying the regularization parameter $\lambda$ from $0.0012$ to $0.01$ for weight decay increases the training accuracy from $55.65\%$ to $97.18\%$. This increase is expected since it corresponds to decreasing the amount of regularization on the weights, i.e., increasing the contribution of the classification error in the optimization objective. However, as it can be seen from Table \ref{table:apx_40} and Figure \ref{fig:40corruption}, the test accuracy behaves rather erratic for weight decay. It can achieve better generalization performance than SGD for small values of $\lambda$, while the performance is very sensitive to the choice of $\lambda$. It is also worth noting that different runs of weight decay with the same hyperparameters often result in (sometimes drastically) different test accuracies.

On the other hand, for RMD, as $\lambda$ varies from $0.7$ to $16.0$, the training accuracy increases from $50.21\%$ to $98.45\%$. This increase is again expected since RMD reduces to SMD as $\lambda\to\infty$, which means that it would interpolate the training data as $\lambda\to\infty$. Unlike weight decay, the test accuracy behaves gracefully and predictably with varying $\lambda$ values. Furthermore, for all $\lambda$ values between $1.0$ and $2.0$, it achieves a test accuracy greater than $80\%$, with the peak of $83.63\%$. This is a whopping $16\%$ improved test accuracy over SGD and $9\%$ improved test accuracy over the best-performing weight decay result. This further highlights the superior generalization performance of RMD via solely solving a different optimization problem and makes it a viable option for training on corrupted datasets. 

Early stopping has been considered as one of the prominent methods to regularize the neural network training \citep{caruana2001overfitting,prechelt1998early}. However, the stopping criterion or the method is often ad-hoc and tailored for each architecture and dataset based on various heuristics, without a mathematical prescription. RMD potentially removes the need for such heuristics and provides a well-formulated and guaranteed stopping criterion, i.e., when the constraints are satisfied. The experiments further show that the points that RMD converges to also generalize well and provide robust test accuracy to small variations in the regularization parameter. We believe these properties combined make RMD a strong candidate to train DNNs.

\begin{table*}[h]
\centering
\caption{Comparison between RMD and the two baselines, (1) implicit regularization induced by SGD and (2) explicit regularization through weight decay, for $40\%$ corruption}
\label{table:apx_40}
 \begin{tabular}{c c  c c }
 \toprule
 \textbf{Algorithm} &  \textbf{Lambda} &    \textbf{Training Accuracy} &   \textbf{Test Accuracy}  \\
  \midrule
  SGD & N/A  &$100\%$ & $67.41\%$ \\
 \midrule
   RMD & $0.7$ & $50.21\%$  & $74.64\%$  \\
   & $0.8$ & $53.36\%$  & $77.88\%$ \\
   & $0.9$ & $55.86\%$  &  $80.81\%$ \\
    & $1.0$ & $57.86\%$  & $83.02\%$ \\
    
     & $1.1$ & $59.26\%$  & $82.99\%$ \\
     
     & $1.2$ & $60.51\%$ & $83.63\%$ \\
      & $1.3$ & $61.70\%$ & $83.49\%$ \\
     & $1.4$ & $62.62\%$&  $83.34\%$ \\
     & $1.5$ & $63.75\%$ & $82.29\%$ \\
      & $1.6$ & $64.42\%$  & $82.58\%$ \\
      & $1.7$ & $65.84\%$ & $82.26\%$ \\
      & $1.8$ & $66.71\%$  & $81.76\%$  \\
      & $1.9$ & $66.62\%$ & $81.03\%$  \\
      & $2.0$ & $68.80\%$ & $81.20\%$ \\
      & $4.0$ & $84.88\%$  & $74.28\%$ \\
     & $16.0$ & $98.45\%$  &$68.80\%$ \\
     
     \midrule
  Weight Decay& $0.0012$ & $55.65\%$  & $74.07\%$ \\
   & $0.0015$ & $60.14\%$ & $73.46\%$ \\
   & $0.0018$ & $75.34\%$ & $54.79\%$ \\
  &$0.002$ & $82.26\%$ & $54.26\%$ \\
  &$0.003$ & $91.46\%$ & $57.19\%$ \\
  &$0.005$ & $95.57\%$  & $59.70\%$\\
  &$0.01$ & $97.18\%$  & $63.00\%$\\
\end{tabular}
\end{table*}

\newpage

\subsection{Results on $25\%$ Corruption of the Training Set}

The training and test accuracies for the three algorithms with various values of $\lambda$ for $25\%$ corruption are summarized in Table~\ref{table:apx_25}. As mentioned before, because the network is highly overparameterized, SGD expectedly interpolates the training data and achieves almost $100\%$ training accuracy. The test accuracy for SGD is just shy of $80\%$. Varying the regularization parameter $\lambda$ for weight decay, it can only achieve marginally better generalization performance than SGD, while often performing worse. 

RMD, on the other hand, significantly outperforms SGD and all the different runs of weight decay, achieving a top test accuracy of almost $87\%$. This performance is not too sensitive to $\lambda$ in the ranges we considered. Moreover, as expected with increasing $\lambda$, it reduces to SGD.

\begin{table*}[h]
\centering
\caption{Comparison between RMD and the two baselines, (1) implicit regularization induced by SGD and (2) explicit regularization through weight decay $25\%$ corruption}

\label{table:apx_25}
 \begin{tabular}{c c  c c }
 \toprule
 \textbf{Algorithm} &  \textbf{Lambda} &    \textbf{Training Accuracy} &   \textbf{Test Accuracy}  \\
  \midrule
  SGD & N/A  &$100\%$ & $79.42\%$ \\
 \midrule
  RMD & $0.7$ & $67.01\%$ & $83.22\%$ \\
   & $0.8$ & $69.15\%$ & $85.14 \%$ \\
   & $0.9$ & $70.20\%$ & $85.43\%$ \\

    & $1.0$ & $71.81\%$ & $85.71\%$ \\
    
     & $1.1$ & $72.90\%$ & $86.12\%$ \\
     
     & $1.2$ & $74.38\%$ & $86.92\%$ \\
      & $1.3$ & $75.04\%$ & $86.35\%$ \\
     & $1.4$ & $75.16\%$ & $86.62\%$ \\
     & $1.5$ & $75.84\%$ & $85.74\%$ \\
      & $1.6$ & $77.56\%$ & $85.72\%$ \\
      & $1.6$ & $78.36\%$ & $85.69\%$ \\
      & $1.7$ & $78.62\%$ & $86.26\%$ \\
      & $1.8$ & $79.07\%$ & $86.00\%$ \\
      & $1.9$ & $80.41\%$ & $85.01\%$ \\
      & $2.0$ & $82.15\%$ & $84.51\%$ \\
      & $4.0$ & $91.35\%$ & $79.38\%$ \\
     & $16.0$ & $98.67\%$ & $79.03\%$ \\
     \midrule
  Weight Decay& $0.001$ & $69.04\%$  & $67.83 \%$ \\
   & $0.0012$ & $71.38\%$ & $80.21\%$  \\
   & $0.0015$ & $78.43\%$ & $73.82\%$ \\
   & $0.0018$ & $82.26\%$ & $75.37\%$ \\
   & $0.002$ & $86.13\%$ & $66.98\%$ \\
   & $0.003$ & $93.64\%$ & $71.81\%$ \\
   & $0.004$ & $97.97\%$ & $74.06\%$ 
\end{tabular}
\end{table*}

\newpage 

\subsection{Results on $10\%$ Corruption of the Training Set}

The training and test accuracies for the three algorithms with various values of $\lambda$ for $10\%$ corruption are summarized in Table~\ref{table:apx_10}. Interpolating the $10\%$ corrupted training data perfectly, SGD achieves test accuracy of $87.87\%$. Note that even though we have used several different regularization parameters $\lambda$ ranging from $0.001$ to $0.006$, which yield different training errors, weight decay does not outperform SGD, indicating that the standard way of explicit regularization may not provide the desired improvement in the generalization performance in certain datasets. Moreover, the erratic behavior of test accuracy with respect to the choice of $\lambda$ is also present for $10\%$ corrupted training set. 

On the other hand, RMD consistently attains well-behaved test accuracy with respect to the choice of regularization parameter $\lambda$. Even though, in this case, the improvement over SGD is small, the test accuracy of RMD is consistently better for $\lambda$ between $1.1$ and $1.9$, with a peak of $88.49\%$ for $\lambda = 1.4$.

\begin{table*}[h]
\centering
\caption{Comparison between RMD and the two baselines, (1) implicit regularization induced by SGD and (2) explicit regularization through weight decay, for $10\%$ corruption}

\label{table:apx_10}
 \begin{tabular}{c c  c c }
 \toprule
 \textbf{Algorithm} &  \textbf{Lambda} &    \textbf{Training Accuracy} &   \textbf{Test Accuracy}  \\
  \midrule
  SGD & N/A  &$100\%$ & $87.87\%$ \\
 \midrule
   RMD & $0.7$ & $81.24\%$ & $86.09\%$ \\
   & $0.8$ & $83.41\%$ & $87.01\%$ \\
   & $0.9$ & $84.23\%$ & $87.14\%$ \\
    & $1.0$ & $85.53\%$ & $87.48\%$ \\
    
     & $1.1$ & $87.08\%$ & $87.89\%$ \\
     
     & $1.2$ & $87.45\%$ & $88.00\%$ \\
      & $1.3$ & $88.68\%$ & $88.20\%$ \\
     & $1.4$ & $89.05\%$ & $88.49\%$ \\ 
     & $1.45$ & $89.46\%$ & $88.26\%$ \\
     & $1.5$ & $89.91\%$ & $87.02\%$ \\
      & $1.6$ & $90.62\%$ & $87.71\%$ \\
      & $1.7$ & $91.43\%$ & $87.97\%$ \\
      & $1.8$ & $91.76\%$ & $87.99\%$ \\
      & $1.9$ & $92.05\%$ & $88.35\%$ \\
      & $2.0$ & $92.72\%$ & $87.73\%$ \\
      & $4.0$ & $96.85\%$ & $86.17\%$ \\
      & $16.0$ & $99.52\%$ & $86.77\%$ \\
     \midrule
  Weight Decay& $0.001$ & $85.54\%$ & $79.77\%$ \\
   & $0.0012$ & $89.13\%$ & $83.06\%$ \\
   & $0.0015$ & $91.87\%$ & $72.36\%$ \\
   & $0.0018$ & $94.25\%$ & $83.10\%$ \\
   & $0.002$ & $94.72\%$ & $80.31\%$ \\
   & $0.003$ & $96.64\%$ & $74.37\%$ \\
   & $0.004$ & $97.38\%$ & $83.69\%$ \\
   & $0.006$ & $98.40\%$ & $82.12\%$ 
\end{tabular}
\end{table*}

\newpage 

\subsection{Results on Uncorrupted Training Set}

As expected, since the data is uncorrupted, interpolating the training data is the right decision in this setting, thus SGD has the best test accuracy. Both RMD and weight decay approaches have higher test accuracy as $\lambda$ increases, with RMD having superior performance. Note that this empirical observation again highlights the fact that as $\lambda\to\infty$, RMD reduces to SMD as shown in Appendix~\ref{apx:SMD_reduction}.

\begin{table*}[hb]
\centering
\caption{Comparison between RMD and the two baselines, (1) implicit regularization induced by SGD and (2) explicit regularization through weight decay, for $0\%$ corruption}
\label{table:apx_0}
 \begin{tabular}{c c  c c }
 \toprule
 \textbf{Algorithm} &  \textbf{Lambda} &    \textbf{Training Accuracy} &   \textbf{Test Accuracy}  \\
  \midrule
  SGD & N/A  &$100\%$ & $92.81\%$ \\
 \midrule
   RMD & $0.7$ & $88.81\%$  & $84.99\%$ \\
   & $0.8$ & $90.79\%$ & $86.71\%$ \\
   & $0.9$ & $92.69\%$ & $87.86\%$ \\
    & $1.0$ & $94.40\%$ & $88.92\%$ \\
    
     & $1.1$ & $95.18\%$ & $88.98\%$ \\
     
     & $1.2$ & $96.68\%$ & $89.23\%$ \\
      & $1.3$ & $97.04\%$ & $90.27\%$ \\
     & $1.4$ & $96.84\%$ & $89.31\%$ \\
     & $1.5$ & $97.73\%$ & $90.27\%$ \\
      & $1.6$ & $97.48\%$ & $90.39\%$ \\
      & $1.7$ & $97.66\%$ & $90.54\%$ \\
      & $1.8$ & $97.43\%$ & $90.77\%$ \\
      & $1.9$ & $97.84\%$ & $90.06\%$ \\
      & $2.0$ & $97.56\%$  & $90.34\%$ \\
      & $4.0$ & $99.45\%$  & $91.0\%$ \\
      & $8.0$ & $99.81\%$  & $91.35\%$ \\
      & $16.0$ & $100\%$  & $92.35\%$  \\
     \midrule
  Weight Decay& $0.001$ & $95.31\%$ & $84.56\%$ \\
  & $0.0012$ & $96.32\%$ & $76.13\%$ \\
   & $0.0015$ & $97.22\%$ & $90.48\%$  \\
   & $0.0018$ & $97.77\%$ & $90.96\%$ \\
  &$0.002$ & $98.08\%$ & $89.01\%$ \\
  &$0.003$ & $98.93\%$ & $ 91.08\%$ \\
  &$0.005$ & $99.29\%$  & $91.62\%$
\end{tabular}
\end{table*}

\end{document}